\documentclass{article}

\usepackage{microtype}
\usepackage{graphicx}
\usepackage{subfigure}
\usepackage{booktabs} 
\usepackage{enumitem}

\usepackage{hyperref}

\usepackage[utf8]{inputenc} 
\usepackage[T1]{fontenc}    
\usepackage{booktabs}       
\usepackage{amsfonts}       
\usepackage{nicefrac}       
\usepackage{microtype}      
\usepackage{multicol}
\usepackage{makecell}
\usepackage{multirow}
\usepackage{siunitx}
\usepackage{wrapfig,lipsum}
\usepackage{caption}
\usepackage{pifont}
\usepackage{graphicx}
\usepackage{wrapfig}
\usepackage{amsmath}
\usepackage{amssymb}
\usepackage[dvipsnames]{xcolor}
\definecolor{mypink}{rgb}{0.858, 0.188, 0.478}
\definecolor{brightpink}{rgb}{1.0, 0.0, 0.5}
\usepackage{color}
\usepackage{multirow}
\usepackage{array} 
\usepackage{tabularx}
\usepackage{booktabs}
\usepackage{caption}

\usepackage{xcolor}

\usepackage[accepted]{icml2021arxiv}

\usepackage{hyperref}       
\usepackage{url}            

\begin{document}

\twocolumn[

\icmltitle{Is Space-Time Attention All You Need for Video Understanding?} 



\icmlsetsymbol{equal}{*}

\begin{icmlauthorlist}
\icmlauthor{Gedas Bertasius}{to}
\icmlauthor{Heng Wang}{to}
\icmlauthor{Lorenzo Torresani}{to,goo}
\end{icmlauthorlist}

\icmlaffiliation{to}{Facebook AI}
\icmlaffiliation{goo}{Dartmouth College}

\icmlcorrespondingauthor{Gedas Bertasius}{gberta@seas.upenn.edu}

\icmlkeywords{Transformers, Action Recognition, Video Classification}

\vskip 0.3in
]



\printAffiliationsAndNotice{}  

\begin{abstract}


We present a convolution-free approach to video classification built exclusively on self-attention over space and time. Our method, named ``TimeSformer,'' adapts the standard Transformer architecture to video by enabling spatiotemporal feature learning directly from a sequence of frame-level patches. Our  experimental study compares different self-attention schemes and suggests that ``divided attention,'' where temporal attention and spatial attention are separately applied within each block, leads to the best video classification accuracy among the design choices considered. Despite the radically new design, TimeSformer achieves state-of-the-art results on several action recognition benchmarks, including the best reported accuracy on Kinetics-400 and Kinetics-600. Finally, compared to 3D convolutional networks, our model is faster to train, it can achieve dramatically higher test efficiency (at a small drop in accuracy), and it can also be applied to much longer video clips (over one minute long). Code and models are available at:  \url{https://github.com/facebookresearch/TimeSformer}.

\end{abstract}
\vspace{-0.3cm}

\section{Introduction}

Over the last few years, the field of natural language processing (NLP) has been revolutionized by the emergence of methods based on self-attention~\cite{NIPS2017_3f5ee243}. Because of their excellent capabilities at capturing long-range dependencies among words as well as their training scalability, self-attention architectures, such as the Transformer model, represent the current state-of-the-art across a wide range of language tasks, including machine translation~\cite{ott-etal-2018-scaling, chen-etal-2018-best}, question answering~\cite{devlin-etal-2019-bert, dai-etal-2019-transformer}, and autoregressive word generation~\cite{radford2019language, brown2020language}.

Video understanding shares several high-level similarities with NLP. First of all, videos and sentences are both sequential. Furthermore, precisely as the meaning of a word can often be understood only by relating it to the other words in the sentence, it may be argued that atomic actions in short-term segments need to be contextualized with the rest of the video in order to be fully disambiguated. Thus, one would expect the long-range self-attention models from NLP to be highly effective for video modeling as well. However, in the video domain, 2D or 3D convolutions still represent the core operators for spatiotemporal feature learning across different video tasks~\cite{Feichtenhofer_2019_ICCV,  DBLP:conf/eccv/TeedD20, gberta_2020_CVPR}. While self-attention has shown benefits when applied on top of convolutional layers~\cite{Wang_2018_CVPR}, to the best of our knowledge, no attempt to use self-attention as the exclusive building block for video recognition models has been reported. 

In this work we pose the question of whether it may be possible to build a performant convolution-free video architecture by replacing altogether the convolution operator with self-attention. We argue that such a design has the potential to overcome a few inherent limitations of convolutional models for video analysis. First, while their strong inductive biases (e.g., local connectivity and translation equivariance) are undoubtedly beneficial on small training sets, they may excessively limit the expressivity of the model in settings where there is ample availability of data and ``all'' can be learned from examples. 
Compared to CNNs, Transformers impose less restrictive inductive biases. This broadens the family of functions they can represent~\cite{Cordonnier:ICLR2020, Zhao:CVPR2020}, and renders them better suited to modern big-data regimes where there is less need for strong inductive priors. Second, while convolutional kernels are specifically designed to capture short-range spatiotemporal information, they cannot model dependencies that extend beyond the receptive field. While deep stacks of convolutions~\cite{Simonyan14c, 43022, DBLP:conf/cvpr/CarreiraZ17} naturally extend the receptive field, these strategies are inherently limited in capturing long-range dependencies by means of aggregation of shorter-range information. Conversely, the self-attention mechanism can be applied to capture both local as well as global long-range dependencies by directly comparing feature activations at all space-time locations, much beyond the receptive field of traditional convolutional filters. Finally, despite the advances in GPU hardware acceleration, training deep CNNs remains very costly, especially when applied to high-resolution and long videos. Recent work in the still-image domain~\cite{Dosovitskiy:ICLR2021, DETR, Zhao:CVPR2020} has demonstrated that Transformers enjoy faster training and inference compared to CNNs,  making it possible to construct models with larger learning capacity for the same computational budget. 

Motivated by these observations, we propose a video architecture built exclusively on self-attention. We adapt the image model ``Vision Transformer'' (ViT)~\cite{Dosovitskiy:ICLR2021} to video by extending the self-attention mechanism from the image space to the space-time 3D volume. Our proposed model, named ``TimeSformer'' (from Time-Space Transformer), views the video as a sequence of patches extracted from the individual frames. As in ViT, each patch is linearly mapped into an embedding and augmented with positional information. This makes it possible to interpret the resulting sequence of vectors as token embeddings which can be fed to a Transformer encoder, analogously to the token features computed from words in NLP. 

One downside of self-attention in standard Transformer is that it requires computing a similarity measure for all pairs of tokens. In our setting, this is computationally costly due to the large number of patches in the video. To address these challenges, we propose several scalable self-attention designs over the space-time volume and empirically evaluate them over large-scale action classification datasets. Among the proposed schemes, we found that the best design is represented by a ``divided attention'' architecture which separately applies temporal attention and spatial attention within each block of the network. Compared to the established paradigm of convolution-based video architecture, TimeSformer follows a radically different design. Yet, it achieves accuracy comparable, and in some cases superior, to the state-of-the-art in this field. We also show that our model can be used for long-range modeling of videos spanning many minutes.

\section{Related Work}


Our approach is influenced by recent works that use self-attention for image classification, either in combination with the convolution operator or even as a full replacement for it. Within the former class, Non-Local Networks~\cite{NLN} employ a non-local mean that effectively generalizes the self-attention function of Transformers~\cite{Vaswani:2017}. Bello et al.~\cite{BelloEtAl} propose a 2D self-attention mechanism that is competitive as a replacement of 2D convolution but gives even stronger results when used to augment convolutional features with self-attention features. Beyond image categorization, Relation Networks~\cite{Hu:CVPR2018} and DETR~\cite{DETR} use self-attention on top of convolutional feature maps for object detection.

Our method is more closely related to image networks leveraging self-attention as a substitute for convolution~\cite{Parmar:ICML2018, Ramachandran:NeurIPS2019, Cordonnier:ICLR2020, Zhao:CVPR2020}. Since these works use individual pixels as queries, in order to maintain a manageable computational cost and a small memory consumption, they must restrict the scope of self-attention to local neighborhoods or use global self-attention on heavily downsized versions of the image. Alternative strategies for scalability to full images include sparse key-value sampling~\cite{Child:OpenAI} or constraining the self-attention to be calculated along the spatial axes~\cite{Ho:Axial, huang2018ccnet, Wang:Axial}. A few of the self-attention operators considered in our experiments adopt similar sparse and axial computation, although generalized to the spatiotemporal volume. However, the efficiency of our approach stems mainly from decomposing the video into a sequence of frame-level patches and then feeding linear embeddings of these patches as input token embeddings to a Transformer. This strategy was recently introduced in Vision Transformers (ViT)~\cite{Dosovitskiy:ICLR2021} which were shown to deliver impressive performance on image categorization. In this work, we build on the ViT design, and extend it to video by proposing and empirically comparing several scalable schemes for space-time self-attention over videos.






While Transformers have been recently used for video generation~\cite{Weissenborn:ICLR2020}, we are not aware of prior video recognition architectures using self-attention as the exclusive building block. However, we note that Transformers have been adopted on top of convolutional feature maps for action localization and recognition~\cite{ActionTransformer}, video classification~\cite{NLN, doubleattention},  and group activity recognition~\cite{Gavrilyuk:CVPR2020}. We also note that there is a wide literature based on the use of text Transformers combined with video CNNs to address various video-language tasks, such as captioning~\cite{zhou2018end}, question-answering~\cite{yang2020bert} and dialog~\cite{le2019multimodal}. Finally, multimodal video-text transformers~\cite{VideoBERT, li2020hero} have also been trained or pretrained in unsupervised fashion by adopting masked-token pretext tasks adapted from the language domain~\cite{BERT,radford2018improving}. 

\begin{figure*}
\begin{center}
   \includegraphics[width=0.8\linewidth]{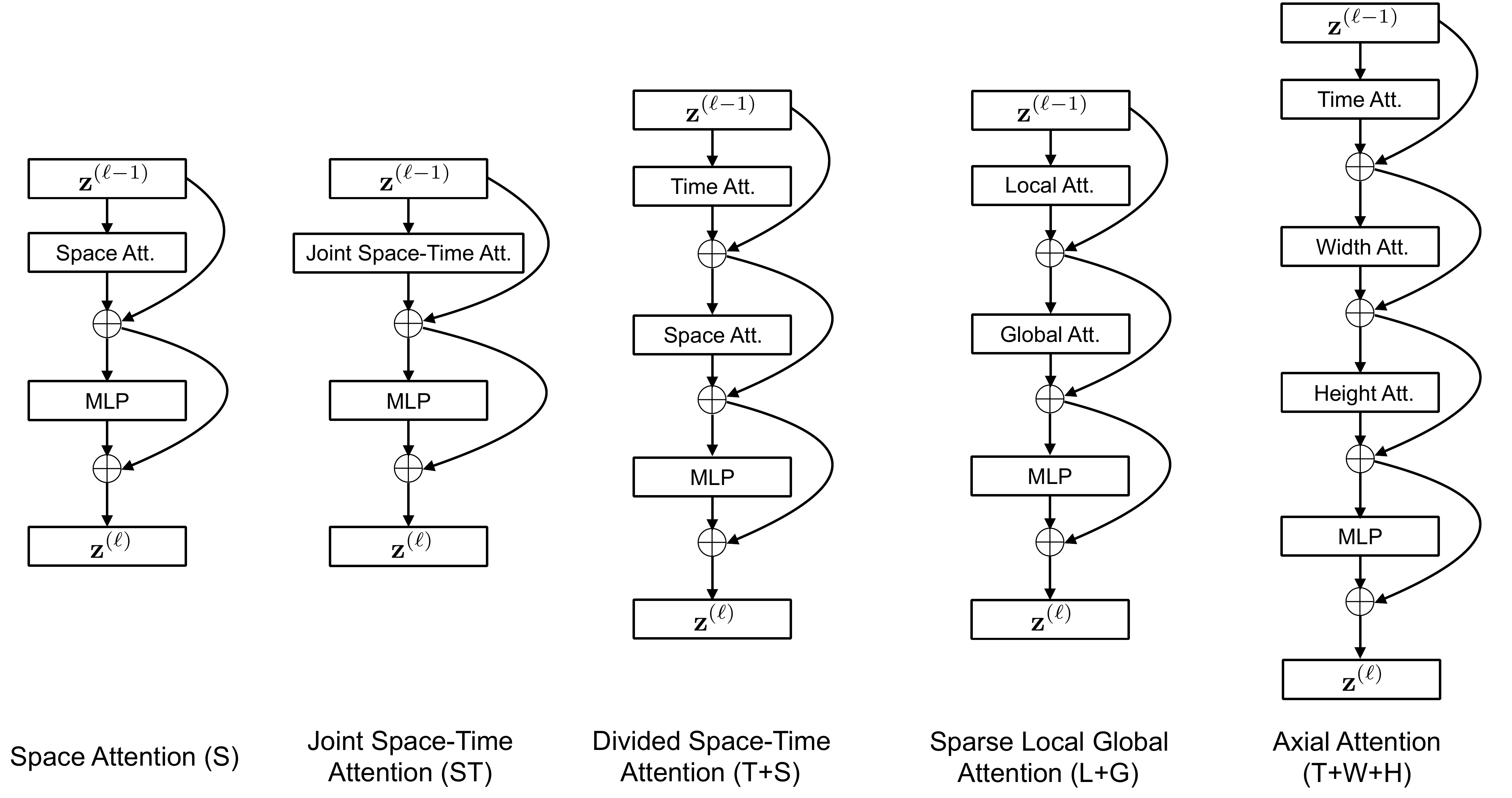}
\end{center}
\vspace{-0.3cm}
        \caption{The video self-attention blocks that we investigate in this work. Each attention layer implements self-attention~\cite{Vaswani:2017} on a specified spatiotemporal neighborhood of frame-level patches (see Figure~\ref{att_fig} for a visualization of the neighborhoods). We use residual connections to aggregate information from different attention layers within each block. A 1-hidden-layer MLP is applied at the end of each block. The final model is constructed by repeatedly stacking these blocks on top of each other.\vspace{-0.1cm}} 
\label{arch_fig}
\end{figure*}

\section{The TimeSformer Model}

\textbf{Input clip.} The TimeSformer takes as input a clip $X \in \mathbb{R}^{H \times W \times 3 \times F}$ consisting of $F$ RGB frames of size $H \times W$ sampled from the original video. 

\textbf{Decomposition into patches.} Following the ViT~\cite{Dosovitskiy:ICLR2021}, we decompose each frame into $N$ non-overlapping patches, each of size $P \times P$, such that the $N$ patches span the entire frame, i.e., $\smash{N=HW/P^2}$. We flatten these patches into vectors $\smash{{\bf x}_{(p,t)} \in \mathbb{R}^{3 P^2}}$ with $p=1,\hdots,N$ denoting spatial locations and $t=1,\hdots,F$ depicting an index over frames.


\textbf{Linear embedding.} We linearly map each patch $\smash{{\bf x}_{(p,t)}}$ into an embedding vector $\smash{{\bf z}_{(p,t)}^{(0)} \in \mathbb{R}^{D}}$ by means of a learnable matrix $\smash{E \in \mathbb{R}^{D  \times 3 P^2}}$:

\vspace{-0.1cm}
\begin{equation}
{\bf z}_{(p,t)}^{(0)} = E {\bf x}_{(p,t)} + {\bf e}_{(p,t)}^{pos}
\end{equation}
\vspace{-0.1cm}

where $\smash{{\bf e}_{(p,t)}^{pos} \in \mathbb{R}^{D}}$ represents a learnable positional embedding added to encode the spatiotemporal position of each patch. The resulting sequence of embedding vectors $\smash{{\bf z}_{(p,t)}^{(0)}}$ for $p=1,\hdots,N$, and $t=1,\hdots,F$ represents the input to the Transformer, and plays a role similar to the sequences of embedded words that are fed to text Transformers in NLP. As in the original BERT Transformer~\cite{BERT}, we add in the first position of the sequence a special learnable vector $\smash{{\bf z}_{(0,0)}^{(0)} \in \mathbb{R}^{D}}$ representing the embedding of the classification token. 

\textbf{Query-Key-Value computation.} Our Transformer consists of $L$ encoding blocks. At each block $\ell$, a query/key/value vector is computed for each patch from the representation $\smash{{\bf z}_{(p,t)}^{(\ell-1)}}$ encoded by the preceding block:

\vspace{-0.4cm}
\begin{align}
{\bf q}_{(p,t)}^{(\ell,a)} &= W_Q^{(\ell,a)} \mathrm{LN}\left({\bf z}_{(p,t)}^{(\ell-1)}\right) \in \mathbb{R}^{D_h}\\
{\bf k}_{(p,t)}^{(\ell,a)} &= W_K^{(\ell,a)} \mathrm{LN}\left({\bf z}_{(p,t)}^{(\ell-1)}\right) \in \mathbb{R}^{D_h}\\
{\bf v}_{(p,t)}^{(\ell,a)} &= W_V^{(\ell,a)} \mathrm{LN}\left({\bf z}_{(p,t)}^{(\ell-1)}\right) \in \mathbb{R}^{D_h}
\end{align}
\vspace{-0.3cm}

where $\mathrm{LN}()$ denotes LayerNorm~\cite{DBLP:journals/corr/BaKH16}, $a = 1, \hdots, \mathcal{A}$ is an index over multiple attention heads and $\mathcal{A}$ denotes the total number of attention heads. The latent dimensionality for each attention head is set to $D_h = D / \mathcal{A}$.

\begin{figure*}[t]
\begin{center}
   \includegraphics[width=0.74\linewidth]{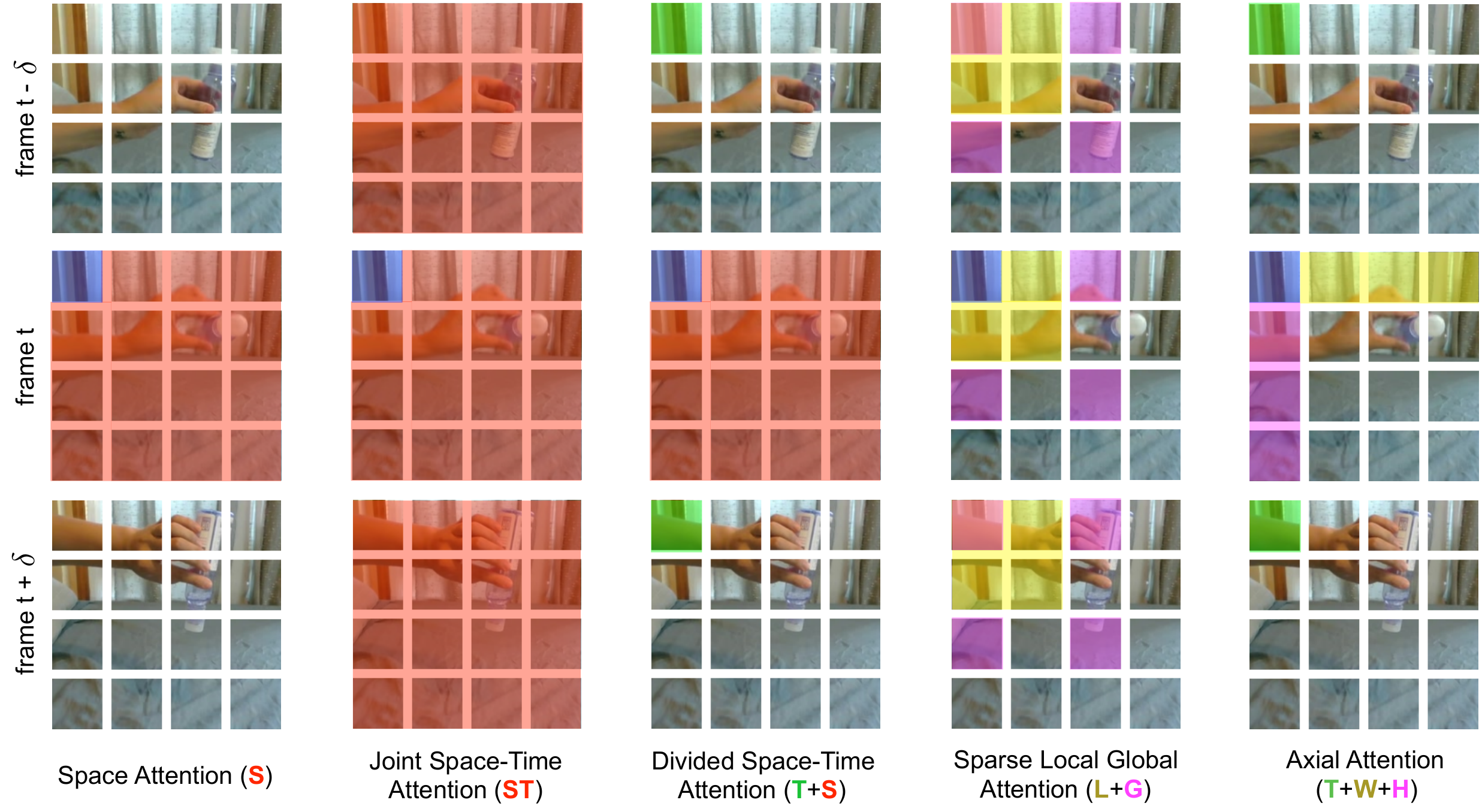}
\end{center}
\vspace{-0.5cm}
        \caption{Visualization of the five space-time self-attention schemes studied in this work. 
        Each video clip is viewed as a sequence of frame-level patches with a size of $16 \times 16$ pixels. For illustration, we denote in blue the query patch and show in non-blue colors its self-attention space-time neighborhood under each scheme. Patches without color are not used for the self-attention computation of the blue patch. Multiple colors within a scheme denote attentions separately applied along different dimensions (e.g., space and time for (T+S)) or over different neighborhoods (e.g., for (L+G)). Note that self-attention is computed for every single patch in the video clip, i.e., every patch serves as a query. We also note that although the attention pattern is shown for only two adjacent frames, it extends in the same fashion to all frames of the clip.\vspace{-0.3cm}}
\label{att_fig}
\end{figure*}

\textbf{Self-attention computation.} Self-attention weights are computed via dot-product. The self-attention weights $\smash{\pmb{\alpha}_{(p,t)}^{(\ell,a)} \in \mathbb{R}^{NF+1}}$ for query patch $(p,t)$ are given by:

\vspace{-0.2cm}
\begin{align}
\pmb{\alpha}_{(p,t)}^{(\ell,a)} &= \mathrm{SM}\left( \frac{{\bf q}_{(p,t)}^{(\ell,a)}}{\sqrt{D_h}}^\top \cdot \left[{\bf k}_{(0,0)}^{(\ell,a)} \left\{ {\bf k}_{(p',t')}^{(\ell,a)} \right\}_{\begin{subarray}{l}p'=1,...,N\\t'=1,...,F
\end{subarray}}  \right] \right)
\label{eq:attentionST}
\end{align}
\vspace{-0.2cm}

where $\mathrm{SM}$ denotes the softmax activation function. Note that when attention is computed over one dimension only (e.g., spatial-only or temporal-only), the computation is significantly reduced. For example, in the case of spatial attention, only $N+1$ query-key comparisons are made, using exclusively keys from the same frame as the query:

\vspace{-0.3cm}
\begin{align}
\pmb{\alpha}_{(p,t)}^{(\ell,a)\mathrm{space}} &= \mathrm{SM}\left( \frac{{\bf q}_{(p,t)}^{(\ell,a)}}{\sqrt{D_h}}^\top \cdot \left[{\bf k}_{(0,0)}^{(\ell,a)} \left\{ {\bf k}_{(p',t)}^{(\ell,a)} \right\}_{p'=1,...,N}  \right] \right).\label{eq:attentionSPACE}
\end{align}
\vspace{-0.3cm}

\textbf{Encoding.} The encoding $\smash{{\bf z}_{(p,t)}^{(\ell)}}$ at block $\ell$ is obtained by first computing the weighted sum of value vectors using self-attention coefficients from each attention head:

\vspace{-0.3cm}
\begin{align}
{\bf s}_{(p,t)}^{(\ell,a)} &= {\alpha}_{(p,t),(0,0)}^{(\ell,a)} {\bf v}_{(0,0)}^{(\ell,a)} + \sum_{p'=1}^N \sum_{t'=1}^F {\alpha}_{(p,t),(p',t')}^{(\ell,a)} {\bf v}_{(p',t')}^{(\ell,a)}.
\end{align}

Then, the concatenation of these vectors from all heads is projected and passed through an MLP, using residual connections after each operation:

\vspace{-0.3cm}
\begin{align}
{\bf z'}_{(p,t)}^{(\ell)} &= W_O \left[ \begin{array}{c} {\bf s}_{(p,t)}^{(\ell,1)}\\ \vdots \\ {\bf s}_{(p,t)}^{(\ell,\mathcal{A})} \end{array} \right] + {\bf z}_{(p,t)}^{(\ell-1)} \label{eq:hiddenstate}\\
{\bf z}_{(p,t)}^{(\ell)} &= \mathrm{MLP}\left(\mathrm{LN}\left({\bf z'}_{(p,t)}^{(\ell)}\right)\right) + {\bf z'}_{(p,t)}^{(\ell)}.\label{eq:encoding}
\end{align}

\textbf{Classification embedding.} The final clip embedding is obtained from the final block for the classification token:

\vspace{-0.1cm}
\begin{equation}
{\bf y} = \mathrm{LN}\left({\bf z}_{(0,0)}^{(L)}\right)  \in  \mathbb{R}^{D}.
\end{equation}
\vspace{-0.2cm}

On top of this representation we append a 1-hidden-layer MLP, which is used to predict the final video classes. 

\textbf{Space-Time Self-Attention Models.} We can reduce the computational cost by replacing the spatiotemporal attention of Eq.~\ref{eq:attentionST} with spatial attention within each frame only (Eq.~\ref{eq:attentionSPACE}). However, such a model neglects to capture temporal dependencies across frames. As shown in our experiments, this approach leads to degraded classification accuracy compared to full spatiotemporal attention, especially on benchmarks where strong temporal modeling is necessary. 

We propose a more efficient architecture for spatiotemporal attention, named ``Divided Space-Time Attention'' (denoted with T+S), where temporal attention and spatial attention are separately applied one after the other. This architecture is compared to that of Space and Joint Space-Time attention in Fig.~\ref{arch_fig}. A visualization of the different attention models on a video example is given in Fig.~\ref{att_fig}. For Divided Attention, within each block $\ell$, we first compute temporal attention by comparing each patch $(p,t)$ with all the patches at the same spatial location in the other frames:

\vspace{-0.3cm}
\begin{align}
\pmb{\alpha}_{(p,t)}^{(\ell,a)\mathrm{time}} &= \mathrm{SM}\left( \frac{{\bf q}_{(p,t)}^{(\ell,a)}}{\sqrt{D_h}}^\top \cdot \left[{\bf k}_{(0,0)}^{(\ell,a)} \left\{ {\bf k}_{(p,t')}^{(\ell,a)} \right\}_{t'=1,...,F}  \right] \right).
\end{align}
\vspace{-0.3cm}

The encoding $\smash{{\bf z'}_{(p,t)}^{(\ell)\mathrm{time}}}$ resulting from the application of Eq.~\ref{eq:hiddenstate} using temporal attention is then fed back for {\em spatial} attention computation instead of being passed to the MLP. In other words, new key/query/value vectors are obtained from ${\bf z'}_{(p,t)}^{(\ell)\mathrm{time}}$ and spatial attention is then computed using Eq.~\ref{eq:attentionSPACE}. Finally, the resulting vector $\smash{{\bf z'}_{(p,t)}^{(\ell)\mathrm{space}}}$ is passed to the MLP of Eq.~\ref{eq:encoding} to compute the final encoding $\smash{{\bf z}_{(p,t)}^{(\ell)}}$ of the patch at block $\ell$. For the model of divided attention, we learn distinct query/key/value matrices $\smash{\{W_{Q^\mathrm{time}}^{(\ell,a)}, W_{K^\mathrm{time}}^{(\ell,a)}, W_{V^\mathrm{time}}^{(\ell,a)}\}}$ and $\smash{\{W_{Q^\mathrm{space}}^{(\ell,a)}, W_{K^\mathrm{space}}^{(\ell,a)}, W_{V^\mathrm{space}}^{(\ell,a)}\}}$ over the time and space dimensions. Note that compared to the $(NF+1)$ comparisons per patch needed by the joint spatiotemporal attention model of Eq.~\ref{eq:attentionST}, Divided Attention performs only $(N + F + 2)$ comparisons per patch. Our experiments demonstrate that this space-time factorization is not only more efficient but it also leads to improved classification accuracy. 

%

\begin{table}[t]
\scriptsize
\begin{center}
\begin{tabular}{ c c   c  c } 
 \hline
 Attention & Params & K400 & SSv2\\ \hline
Space & 85.9M &  76.9 & 36.6 \\
Joint Space-Time & 85.9M & 77.4 & 58.5 \\
Divided Space-Time & 121.4M &  \bf 78.0 & \bf 59.5\\
Sparse Local Global & 121.4M &  75.9 & 56.3 \\
Axial & 156.8M & 73.5 & 56.2 \\
\hline
\end{tabular}
\end{center}
\vspace{-0.2cm}
\caption{Video-level accuracy for different space-time attention schemes in TimeSformer. 
We evaluate the models on the validation sets of Kinetics-400 (K400), and Something-Something-V2 (SSv2). We observe that divided space-time attention achieves the best results on both datasets.\vspace{-0.3cm}} 
\label{att_results_table}
\end{table}

We have also experimented with a ``Sparse Local Global'' (L+G) and an ``Axial'' (T+W+H) attention models. Their architectures are illustrated in Fig.~\ref{arch_fig}, while Fig.~\ref{att_fig} shows the patches considered for attention by these models. For each patch $(p,t)$, (L+G) first computes a local attention by considering the neighboring $F \times H/2 \times W/2$ patches and then calculates a sparse global attention over the entire clip using a stride of 2 patches along the temporal dimension and also the two spatial dimensions. Thus, it can be viewed as a faster approximation of full spatiotemporal attention using a local-global decomposition and a sparsity pattern, similar to that used in~\cite{Child:OpenAI}. Finally, ``Axial'' attention decomposes the attention computation in three distinct steps: over time, width and height. A decomposed attention over the two spatial axes of the image was proposed in~\cite{Ho:Axial, huang2018ccnet, Wang:Axial} and our (T+W+H) adds a third dimension (time) for the case of video. All these models are implemented by learning distinct query/key/value matrices for each attention step. 

\section{Experiments}



We evaluate TimeSformer on four popular action recognition datasets: Kinetics-400~\cite{DBLP:conf/cvpr/CarreiraZ17}, Kinetics-600~\cite{DBLP:journals/corr/abs-1808-01340}, Something-Something-V2~\cite{DBLP:journals/corr/GoyalKMMWKHFYMH17}, and Diving-48~\cite{Li_2018_ECCV}. We adopt the ``Base'' ViT architecture~\cite{Dosovitskiy:ICLR2021} pretrained on either ImageNet-1K or ImageNet-21K~\cite{5206848}, as specified for each experiment. Unless  differently indicated, we use clips of size $8 \times 224 \times 224$, with frames sampled at a rate of $1/32$. The patch size is $16\times16$ pixels. During inference, 
unless otherwise noted, 
we sample a single temporal clip in the middle of the video. We use $3$ spatial crops (top-left, center, bottom-right) from the temporal clip and obtain the final prediction by averaging the scores for these $3$ crops. 



\begin{figure}
\begin{center}
   \includegraphics[width=0.92\linewidth]{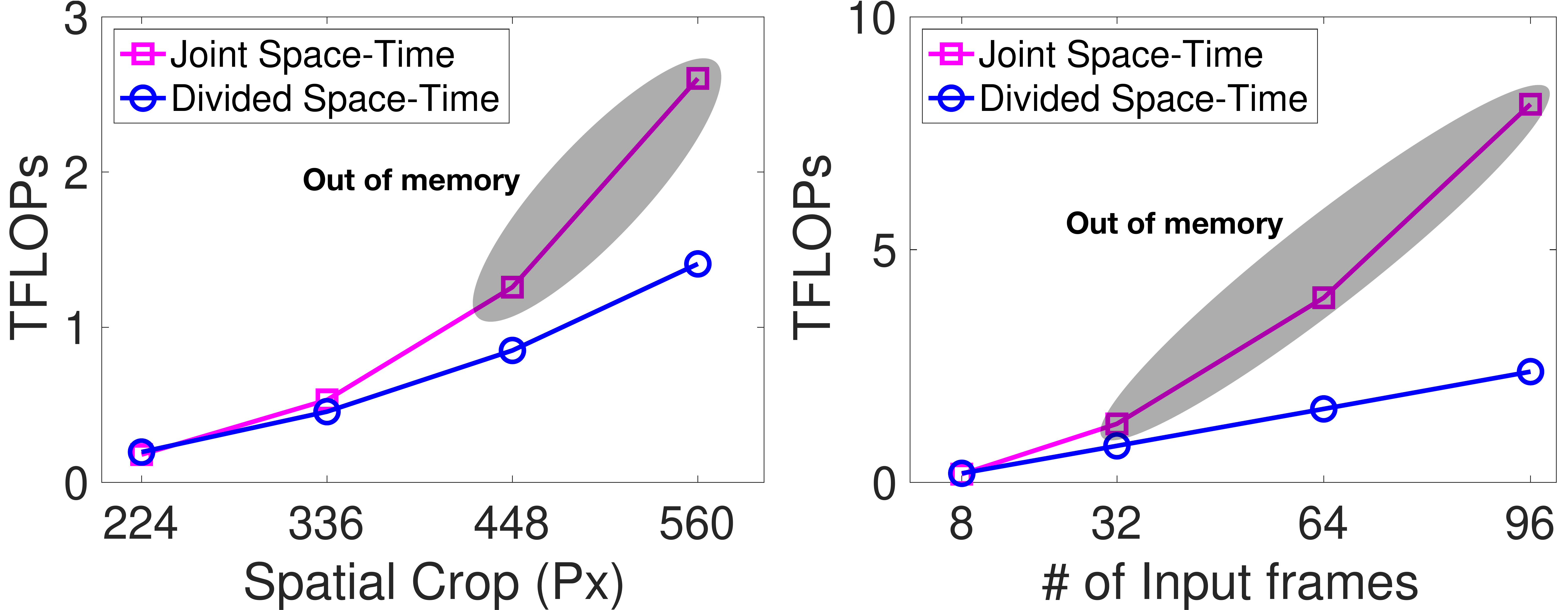}
\end{center}
\vspace{-0.4cm}
        \caption{We compare the video classification cost (in TFLOPs) of Joint Space-Time versus Divided Space-Time attention. We plot the number of TFLOPs as a function of spatial crop size in pixels (left), and the number of input frames (right). As we increase the spatial resolution (left), or the video length (right), our proposed divided space-time attention leads to dramatic computational savings compared to the scheme of joint space-time attention. \vspace{-0.4cm}}
\label{cost_comparison_fig}
\end{figure}

\subsection{Analysis of Self-Attention Schemes}
\label{attn_sec}

For this first set of experiments we start from a ViT pretrained on ImageNet-21K. In Table~\ref{att_results_table}, we present the results obtained with TimeSformer for the five proposed space-time attention schemes on Kinetics-400 (K400) and Something-Something-V2 (SSv2). First, we note that TimeSformer with space-only attention (S) performs well on K400. This is an interesting finding. Indeed, prior work~\cite{Sevilla-Lara_2021_WACV} has shown that on K400, spatial cues are more important than temporal information in order to achieve strong accuracy. Here, we show that it is possible to obtain solid accuracy on K400 without any temporal modeling. Note, however, that space-only attention performs poorly on SSv2. This stresses the importance of temporal modeling on this latter dataset.




Furthermore, we observe that divided space-time attention achieves the best accuracy on both K400 and SSv2. This makes sense because compared to joint space-time attention, divided space-time attention has a larger learning capacity (see Table~\ref{att_results_table}) as it contains distinct learning parameters for temporal attention and spatial attention. 

In Figure~\ref{cost_comparison_fig}, we also compare the computational cost of joint space-time versus divided space-time attention when using higher spatial resolution (left) and longer (right) videos. We note that the scheme of divided space-time scales gracefully under both of these settings. In contrast, the scheme of joint space-time attention leads to a dramatically higher cost when resolution or video length is increased. In practice, joint space-time attention causes a GPU memory overflow once the spatial frame resolution reaches $448$ pixels, or once the number of frames is increased to $32$ and thus it is effectively not applicable to large frames or long videos. Thus, despite a larger number of parameters, divided space-time attention is more efficient than joint space-time attention when operating on higher spatial resolution, or longer videos. Thus, for all subsequent experiments we use a TimeSformer constructed with divided space-time self-attention blocks.






 \begin{table}
\centering
\setlength{\tabcolsep}{3pt}
{\scriptsize
 \begin{tabular}{c c c c c c }
 \hline
{Model} & {Pretrain} & \multicolumn{1}{p{1.7cm}}{\centering K400 Training \\ Time (hours)} & \multicolumn{1}{p{0.6cm}}{\centering K400 \\ Acc.}&  \multicolumn{1}{p{1cm}}{\centering Inference \\  TFLOPs}  & Params\\  
 \hline
I3D 8x8 R50 &  ImageNet-1K & 444 & 71.0 & 1.11 & 28.0M\\ 
I3D 8x8 R50 &  ImageNet-1K & 1440 & 73.4  & 1.11 & 28.0M\\ \hline
SlowFast R50  & ImageNet-1K & 448 & 70.0  & 1.97 & 34.6M\\
SlowFast R50  & ImageNet-1K &  3840 & 75.6 & 1.97 & 34.6M\\
SlowFast R50 & N/A & 6336 & 76.4 & 1.97 & 34.6M\\ \hline
TimeSformer & ImageNet-1K & \bf 416  & 75.8 & \bf 0.59 & 121.4M\\
TimeSformer   & ImageNet-21K & \bf 416  & \bf 78.0 & \bf 0.59 & 121.4M\\
 \hline
 \end{tabular}
 }
\vspace{-0.3cm}
\caption{Comparing TimeSformer to SlowFast and I3D. We observe that TimeSformer has lower inference cost despite having a larger number of parameters. Furthermore, the cost of training TimeSformer on video data is much lower compared to SlowFast and I3D, even when all models are pretrained on ImageNet-1K.\vspace{-0.3cm}} 
\label{comparison_table}
 \end{table}

\subsection{Comparison to 3D CNNs}
\label{comparison_sec}

In this subsection we perform an empirical study aimed at understanding the distinguishing properties of TimeSformer compared to 3D convolutional architectures, which have been the prominent approach to video understanding in recent years. We focus our comparison on two 3D CNN models: 1) SlowFast~\cite{slowfast}, which is the state-of-the-art in video classification, and 2) I3D~\cite{DBLP:conf/cvpr/CarreiraZ17}, which has been shown to benefit from image-based pretraining, similarly to our own model. We present quantitative comparisons to these two networks in Table~\ref{comparison_table} and highlight key observations below.

\textbf{Model Capacity.} From Table~\ref{comparison_table}, we first observe that although TimeSformer has a large learning capacity (the number of parameters is $121.4M$), it has low inference cost ($0.59$ in TFLOPs).  In contrast, SlowFast 8x8 R50 has a larger inference cost ($1.97$ TFLOPs) despite containing only $34.6M$ parameters. Similarly, I3D 8x8 R50 also has a larger inference cost ($1.11$ TFLOPs) despite containing fewer parameters ($28.0M$). This suggests that TimeSformer is better suited for settings that involve large-scale learning. In contrast, the large computational cost of modern 3D CNNs makes it difficult to further increase their model capacity while also maintaining efficiency.

\textbf{Video Training Time.} One significant advantage of ImageNet pretraining is that it enables very efficient training of TimeSformer on video data. Conversely, state-of-the-art 3D CNNs are much more expensive to train even if pretrained on image datasets. In Table~\ref{comparison_table}, we compare the video training time on Kinetics-400 (in Tesla V100 GPU hours) of TimeSformer to that of SlowFast and I3D. Starting from a ResNet50 pretrained on ImageNet-1K, SlowFast $8\times8$ R50 requires $3\,840$ Tesla V100 GPU hours in order to reach an accuracy of $75.6\%$ on Kinetics-400. Training I3D, under similar settings, requires $1\,440$ Tesla V100 GPU hours for a $73.4\%$ accuracy. In contrast, TimeSformer, also pretrained on ImageNet-1K, only requires $416$ Tesla V100 GPU hours to achieve a higher $75.8\%$ accuracy (see Table~\ref{comparison_table}). Furthermore, if we constrain SlowFast to be trained under a somewhat similar computational budget as TimeSformer (i.e., $448$ GPU hours), its accuracy drops to $70.0\%$. Similarly, training I3D using a similar computational budget (i.e., $444$ GPU hours) leads to a lower accuracy of $71.0\%$. This highlights the fact that some of the latest 3D CNNs~\cite{slowfast,feichtenhofer2020x3d} require a very long optimization schedule to achieve good performance (even when using ImageNet pretraining). In contrast, TimeSformer provides a more efficient alternative to labs that do not have access to hundreds of GPUs.


 \begin{table}
\centering
{
\scriptsize
 \begin{tabular}{c c c c }
 \hline
 {Method} & {Pretraining} & {K400} & {SSv2} \\ 
 \hline
TimeSformer & ImageNet-1K & 75.8  & \bf 59.5 \\
TimeSformer & ImageNet-21K & \bf 78.0  & \bf 59.5 \\ \hline
TimeSformer-HR & ImageNet-1K  & 77.8  &  62.2 \\
TimeSformer-HR & ImageNet-21K  & \bf 79.7  & \bf 62.5 \\ \hline
TimeSformer-L & ImageNet-1K & 78.1 & \bf 62.4\\ 
TimeSformer-L & ImageNet-21K & \bf 80.7 & 62.3\\ 
 \hline
 \end{tabular}
 }
  \vspace{-0.1cm}
\caption{Comparing the effectiveness of ImageNet-1K and ImageNet-21K pretraining on Kinetics-400 (K400) and Something-Something-V2 (SSv2). On K400, ImageNet-21K pretraining leads consistently to a better performance compared to ImageNet-1K pretraining. On SSv2, ImageNet-1K and ImageNet-21K pretrainings lead to similar accuracy.\vspace{-0.3cm}}
\label{pretraining_results_table}
 \end{table}


\textbf{The Importance of Pretraining.} Due to a large number of parameters, training our model from scratch is difficult. Thus, before training TimeSformer on video data, we initialize it with weights learned from ImageNet. In contrast, SlowFast can be learned on video data from scratch although at the expense of a very high training cost  (see Table~\ref{comparison_table}). We also attempted to train TimeSformer on Kinetics-400 directly, without any ImageNet pretraining. By using a longer training schedule and more data augmentations, we found it possible to train the model from scratch, albeit to a much lower video level accuracy of $64.8\%$. Thus, based on these results, for all subsequent studies we continued to use ImageNet for pretraining~\cite{5206848}

In Table~\ref{pretraining_results_table} we study the benefits of ImageNet-1K vs ImageNet-21K pretraining on K400 and SSv2. For these experiments, we use three variants of our model: (1) \textbf{TimeSformer}, which is the default version of our model operating on $8 \times 224 \times 224$ video clips, (2) \textbf{TimeSformer-HR}, a high spatial resolution variant that operates on $16 \times 448 \times 448$ video clips, and lastly (3) \textbf{TimeSformer-L}, a long-range configuration of our model that operates on  $96 \times 224 \times 224$ video clips with frames sampled at a rate of $1/4$.

Based on the results in Table~\ref{pretraining_results_table}, we observe that ImageNet-21K pretraining is beneficial for K400, where it leads to a consistently higher accuracy compared to ImageNet-1K pretraining.  On the other hand, on SSv2, we observe that ImageNet-1K and ImageNet-21K pretrainings lead to similar accuracy. This makes sense as SSv2 requires complex spatiotemporal reasoning, whereas K400 is biased more towards spatial scene information, and thus, it benefits more from the features learned on the larger pretraining dataset.

\begin{figure}
\begin{center}
   \includegraphics[width=0.95\linewidth]{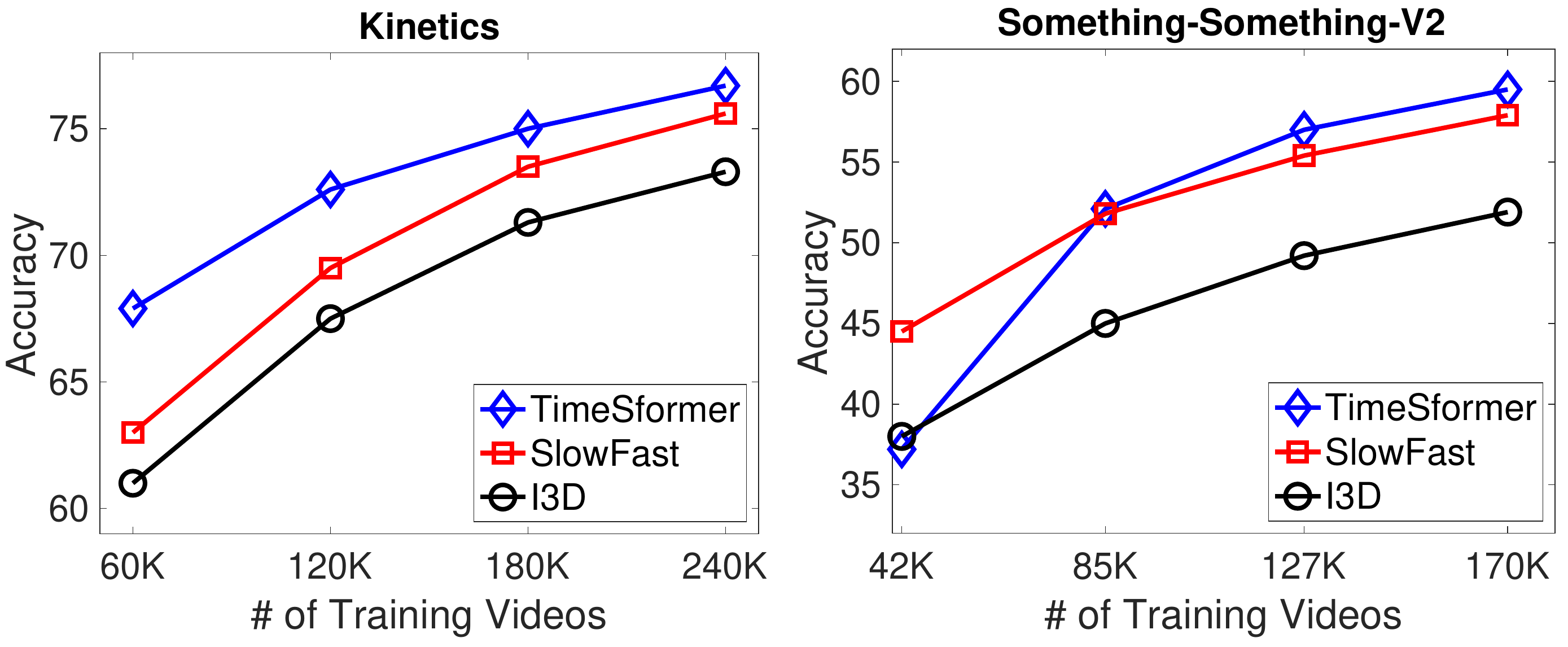}
\end{center}
\vspace{-0.5cm}
        \caption{Accuracy on Kinetics-400 (K400), and Something-Something-V2 (SSv2) as a function of the number of training videos. On K400, TimeSformer performs best in all cases. On SSv2, which requires more complex temporal reasoning, TimeSformer outperforms the other models only when using enough training videos. All models are pretrained on ImageNet-1K.\vspace{-0.2cm}}
\label{data_efficiency_fig}
\end{figure}

\textbf{The Impact of Video-Data Scale.} To understand the effects of video-data scale on performance, we trained TimeSformer on different subsets of K400 and SSv2: $\{25\%, 50\%, 75\%, 100\%\}$ of the full datasets. We show these results in Figure~\ref{data_efficiency_fig}, where we also compare our method with SlowFast R50~\cite{slowfast}, and I3D R50~\cite{DBLP:conf/cvpr/CarreiraZ17} trained on the same subsets and using the same pretraining. Since we do not have access to a ResNet pretrained on ImageNet-21K, we use ImageNet-1K pretraining for all $3$ architectures.


The results of Figure~\ref{data_efficiency_fig} show that, on K400, TimeSformer outperforms the other models for all training subsets. However, we observe a different trend on SSv2, where TimeSformer is the strongest model only when trained on $75\%$ or $100\%$ of the full data. This may be explained by the fact that compared to K400, SSv2 requires learning more complex temporal patterns, and thus more examples are needed by TimeSformer to learn effectively those patterns.

\subsection{Varying the Number of Tokens}
\label{abl_sec}

The scalability of our model allows it to operate at higher spatial resolution and on longer videos compared to most 3D CNNs. We note that both of these aspects affect the length of the sequence of tokens fed to the Transformer. Specifically, increasing the spatial resolution results in a higher number of patches ($N$) per frame. The number of input tokens is also increased when using more frames. To investigate the benefits, we conduct an empirical study where we separately increase the number of tokens along each of these two axes.

\begin{figure}
\begin{center}
   \includegraphics[width=0.92\linewidth]{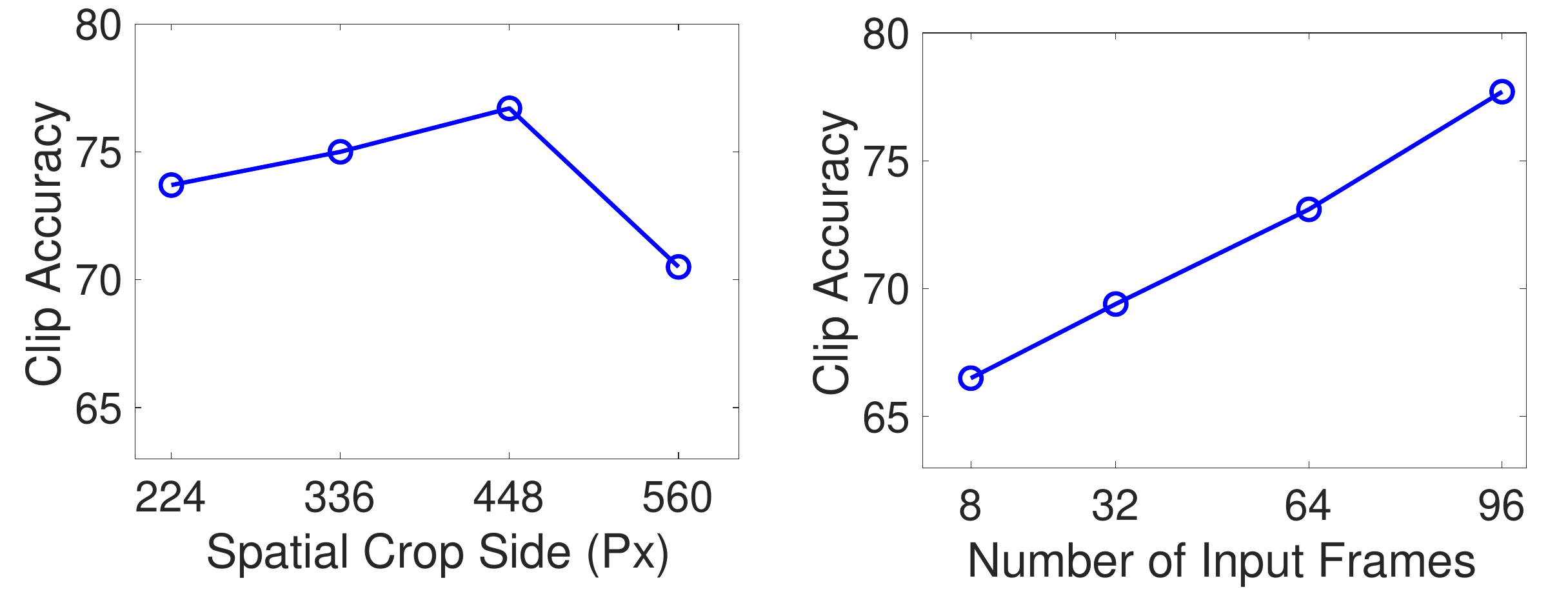}
\end{center}
\vspace{-0.4cm}
        \caption{Clip-level accuracy on Kinetics-400 as a function of spatial crop size in pixels (left), and the number of input frames (right). \vspace{-0.3cm}}
\label{abl_fig}
\end{figure}

We report the findings in Figure~\ref{abl_fig}. We see that increasing the spatial resolution (up to a certain point) leads to a boost in performance. Similarly, we observe that increasing the length of the input clip leads to consistent accuracy gains. Due to GPU memory constraints, we are not able to test our model on clips longer than $96$ frames. Still, we would like to point out that using clips of $96$ frames is a significant departure from current convolutional models, which are typically limited to processing inputs of $8-32$ frames.





 \begin{table}
\centering
{\scriptsize
 \begin{tabular}{c c c }
 \hline
 {Positional Embedding} & {K400} & {SSv2} \\ 
 \hline
None & 75.4  & 45.8 \\
Space-only & 77.8  & 52.5 \\
Space-Time & \bf 78.0 & \bf 59.5\\
 \hline
 \end{tabular}
 }
  \vspace{-0.1cm}
\caption{Ablation on positional embeddings. The version of TimeSformer using space-time positional embeddings yields the highest accuracy on both Kinetics-400 and SSv2.\vspace{-0.3cm}}
\label{pos_results_table}
 \end{table}

%
%



\subsection{The Importance of Positional Embeddings} 



To investigate the importance of our learned spatiotemporal positional embeddings, we also conduct experiments with a few variants of TimeSformer that use: (1) no positional embedding, (2) space-only positional embedding, and (3) space-time positional embedding. We report these results in Table~\ref{pos_results_table}. Based on these results, we observe that the variant of our model that uses space-time positional embeddings produces the best accuracy on both Kinetics-400, and Something-Something-V2. Interestingly, we also observe that using space-only positional embeddings leads to solid results on Kinetics-400, but much worse results on Something-Something-V2. This makes sense as Kinetics-400 is more spatially biased, whereas Something-Something-V2 requires complex temporal reasoning.

 \begin{table}
\footnotesize
\setlength{\tabcolsep}{3pt}
\scriptsize
\begin{center}
 \begin{tabular}{c c c c }
 \hline
 {Method} & {Top-1} & {Top-5} & {TFLOPs} \\ 
 \hline
R(2+1)D~\cite{DBLP:journals/corr/abs-1711-11248} & 72.0 & 90.0 & 17.5\\
bLVNet~\cite{NEURIPS2019_3d779cae} & 73.5 & 91.2 & 0.84\\
TSM~\cite{lin2019tsm} &74.7 & N/A & N/A\\
S3D-G~\cite{DBLP:conf/eccv/XieSHTM18} & 74.7 & 93.4 & N/A \\
Oct-I3D+NL~\cite{Chen_2019_ICCV} & 75.7 & N/A & 0.84\\
D3D~\cite{Stroud_2020_WACV} & 75.9 & N/A & N/A \\
I3D+NL~\cite{NLN} & 77.7 & 93.3 & 10.8 \\
ip-CSN-152~\cite{tran2019video} & 77.8 & 92.8 & 3.2 \\ 
CorrNet~\cite{Wang_2020_CVPR} & 79.2 & N/A & 6.7 \\
LGD-3D-101~\cite{qiu2019learning} & 79.4 & 94.4 & N/A\\ 
SlowFast~\cite{slowfast} & 79.8 & 93.9 & 7.0 \\
X3D-XXL~\cite{feichtenhofer2020x3d} &80.4 & 94.6 & 5.8 \\
\hline
   TimeSformer & 78.0 & 93.7 & \bf 0.59 \\
  TimeSformer-HR  &  79.7 & 94.4 & 5.11 \\
 TimeSformer-L &\bf 80.7 & \bf 94.7 & 7.14 \\
 \hline
 \end{tabular}
 \end{center}
\vspace{-0.3cm}
\caption{{Video-level accuracy on Kinetics-400}. 
\vspace{-0.1cm}}
\label{k400_results_table}
 \end{table}








\subsection{Comparison to the State-of-the-Art}
\label{sota_sec}

\textbf{Kinetics-400 \& Kinetics-600.} In Table~\ref{k400_results_table} we present our results on the validation set of K400. For these experiments, we use TimeSformer pretrained on ImageNet-21K. In addition to the accuracy metrics, we also include inference cost, given in TFLOPs. We note that whereas most previous methods use $10$ temporal clips with $3$ spatial crops (for a total of $30$ space-time views) during inference, TimeSformer achieves solid accuracy with only $3$ views (3 spatial crops), which reduces the inference cost. Our long-range variant, TimeSformer-L achieves a top-1 accuracy of $80.7\%$. Furthermore, our default TimeSformer has the lowest inference cost among recent state-of-the-art models. Yet, it still provides a solid accuracy of $78.0\%$, outperforming many more costly models. 

We also measured the actual inference runtime on $20K$ validation videos of Kinetics-400 (using $8$ Tesla V100 GPUs). Whereas SlowFast takes $14.88$ hours to complete the inference, TimeSformer, TimeSformer-HR, and TimeSformer-L take $36$ minutes, $1.06$ hours and $2.6$ hours, respectively. Thus, even though SlowFast and TimeSformer-L have comparable cost in terms of TFLOPs, in practice the runtimes of all our versions of TimeSformer are much lower. 

In Table~\ref{k600_results_table}, we also present our results on Kinetics-600. Just like on Kinetics-400, we observe that TimeSformer performs well on this benchmark, outperforming all prior methods.




 \begin{table}
\centering
{\scriptsize
 \begin{tabular}{c c c }
 \hline
 {Method} & {Top-1} & {Top-5} \\ 
 \hline
I3D-R50+Cell~\cite{DBLP:conf/eccv/WangXNPRAKH20} & 79.8 & 94.4\\
LGD-3D-101~\cite{qiu2019learning} & 81.5 & 95.6\\ 
SlowFast~\cite{slowfast} & 81.8 & 95.1\\
X3D-XL~\cite{feichtenhofer2020x3d} & 81.9 & 95.5\\
\hline
TimeSformer & 79.1 & 94.4\\
TimeSformer-HR & 81.8 & \bf 95.8\\
TimeSformer-L & \bf 82.2 & 95.6\\
 \hline
 \end{tabular}
 }
\vspace{-0.1cm}
\caption{{Video-level accuracy on Kinetics-600}.\vspace{-0.1cm}} 
\label{k600_results_table}
 \end{table}

\begin{figure}
\begin{center}
   \includegraphics[width=0.6\linewidth]{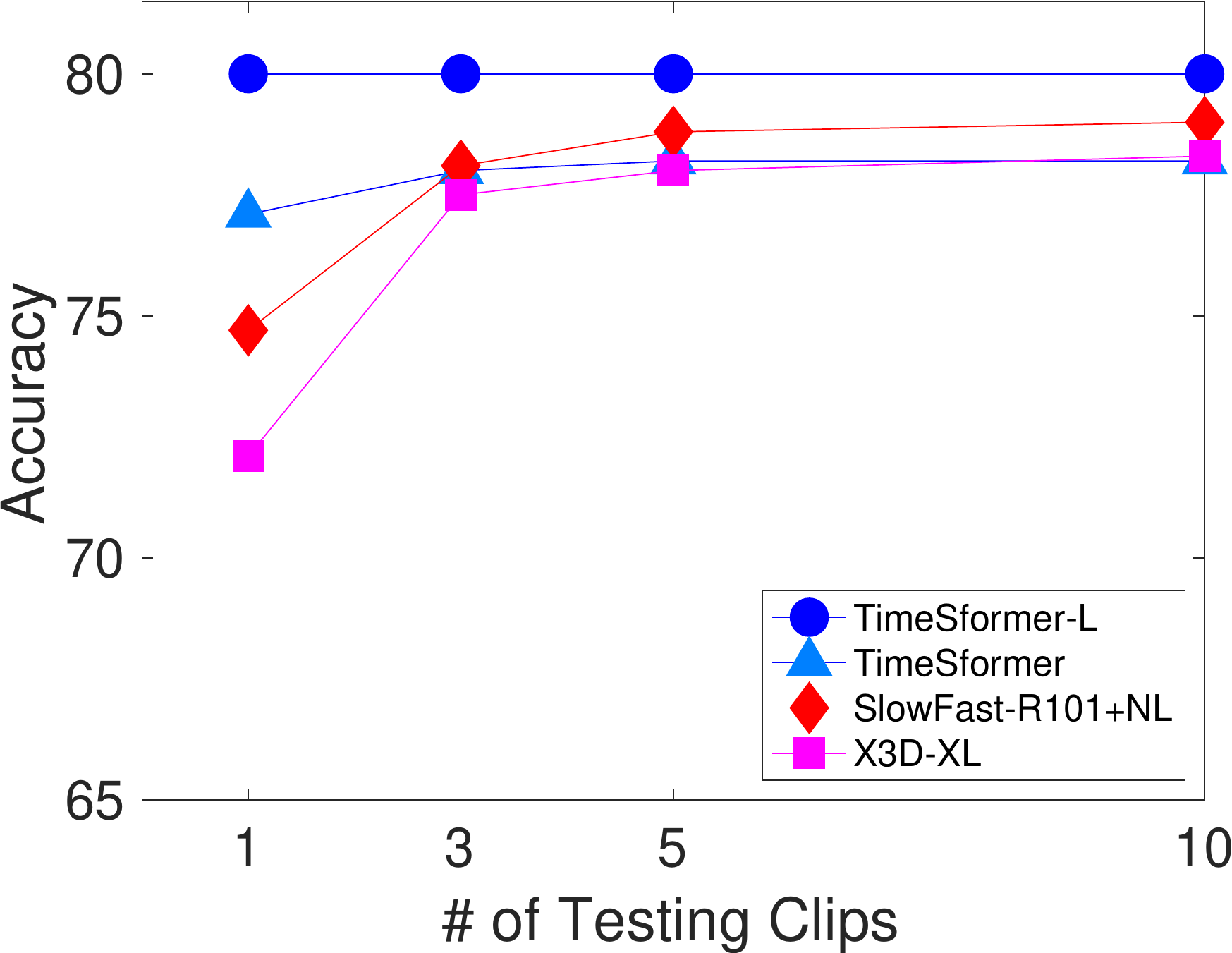}
\end{center}
\vspace{-0.4cm}
        \caption{Video-level accuracy on Kinetics-400 vs the number of temporal clips used during inference. TimeSformer-L achieves excellent accuracy using a small number of clips, which leads to strong performance at low inference cost.\vspace{-0.0cm}}
\label{inference_cost_fig}
\end{figure}

Finally, in Figure~\ref{inference_cost_fig}, we study the effect of using multiple temporal clips during inference (each with a single spatial crop). We plot accuracy using $K \in \{1, 3, 5, 10\}$ temporal clips for testing. We compare our model against X3D~\cite{feichtenhofer2020x3d}, and SlowFast~\cite{slowfast}. X3D and SlowFast require multiple ($\geq 5$) clips to approach their top accuracy. Conversely, our long-range variant, TimeSformer-L, does not require multiple clips to achieve its best performance, since it is able to span about $12$ seconds of a Kinetics video with a single clip.


\textbf{Something-Something-V2 \& Diving-48.} In Table~\ref{ssv2_results_table}, we also validate our model on SSv2 and Diving-48. Since ImageNet-21K pretraining does not improve accuracy on SSv2 (see Table~\ref{pretraining_results_table}), in this case, we use TimeSformer pretrained on ImageNet-1K. This also allows us to apply the same pretraining to all other models in this comparison, using a ResNet pretrained on ImageNet-1K. Our results suggest that TimeSformer achieves lower accuracy than the best models on this dataset. However, considering that our model uses a completely different design, we take these results as suggesting that TimesSformer is a promising approach even for challenging temporally-heavy datasets, such as SSv2. 

In Table~\ref{ssv2_results_table}, we also present our method on another ``temporally-heavy'' dataset, Diving-48. Due to a recently discovered issue with a previous version of Diving-48 labels, here, we only compare our method with a reproduced SlowFast $16\times8$ R101 model. Our results show that TimeSformer outperforms SlowFast by a substantial margin.



\begin{table}
\centering
{\scriptsize
 \begin{tabular}{c c c }
 \hline
 {Method} & {SSv2} & {Diving-48\boldmath{$^{**}$}} \\ 
 \hline
SlowFast~\cite{slowfast} & 61.7 & 77.6 \\
TSM~\cite{lin2019tsm} & 63.4 & N/A \\
STM~\cite{Jiang_2019_ICCV}  & 64.2 & N/A \\
MSNet~\cite{kwon2020motionsqueeze} & 64.7 & N/A \\
TEA~\cite{Li_2020_CVPR} & 65.1 & N/A \\
bLVNet~\cite{NEURIPS2019_3d779cae} & \bf 65.2 & N/A \\
\hline
TimeSformer & 59.5  & 74.9 \\
TimeSformer-HR & 62.2  & 78.0 \\
TimeSformer-L & 62.4 & \bf 81.0\\
 \hline
 \end{tabular}
 }
\vspace{-0.1cm}
\caption{{Video-level accuracy on Something-Something-V2 and Diving-48}. 
\boldmath{${^{**}}$}Due to an issue with Diving-48 labels used in previously published results, we only compare our method with a reproduced SlowFast $16 \times 8$ R101 model. All models are pretained on ImageNet-1K.\vspace{-0.3cm}}
\label{ssv2_results_table}
 \end{table}

\subsection{Long-Term Video Modeling}
\label{long_sec}

Lastly, we evaluate TimeSformer on the task of long-term video modeling using HowTo100M~\cite{miech19howto100m}. HowTo100M is an instructional video dataset that contains around 1M instructional Web videos showing humans performing over 23K different tasks, such as cooking, repairing, making arts, etc. The average duration of these videos is around $7$ minutes, which is orders of magnitude longer than the duration of videos in standard action recognition benchmarks. Each HowTo100M video has a label indicating the task demonstrated in the video (one out of the 23K classes), which can be used for supervised training. Thus, it is a good benchmark to assess the ability of a model to recognize activities exhibited over very long temporal extents.

For this evaluation, we consider only categories that have at least $100$ video examples. This gives a subset of HowTo100M corresponding to $120K$ videos spanning $1059$ task categories. We randomly partition this collection into $85K$ training videos and $35K$ testing videos. 


\begin{table}[t]
\setlength{\tabcolsep}{3pt}
\scriptsize
\begin{center}
\begin{tabular}{ c c c  c  c } 
 \hline
 Method & \multicolumn{1}{p{1cm}}{\centering \# Input \\ Frames} & \multicolumn{1}{p{1.5cm}}{\centering Single Clip \\ Coverage} & \multicolumn{1}{p{1cm}}{\centering \# Test \\ Clips}   &  \multicolumn{1}{p{0.8cm}}{\centering Top-1 \\ Acc}\\ \hline
 SlowFast & 8 & 8.5s & 48 & 48.2 \\
 SlowFast & 32 & 34.1s & 12 & 50.8 \\ 
  SlowFast & 64 & 68.3s & 6 & 51.5  \\ 
  SlowFast & 96 & 102.4s & 4 &  51.2\\ \hline
 TimeSformer & 8 & 8.5s & 48 & 56.8  \\
 TimeSformer & 32 & 34.1s & 12 & 61.2 \\
 TimeSformer & 64 &  68.3s & 6 & 62.2 \\ 
 TimeSformer & 96 & 102.4s & 4 & \bf 62.6  \\
\hline
\end{tabular}
\end{center}
\vspace{-0.3cm}
\caption{{Long-term task classification on HowTo100M.} Given a video spanning several minutes, the goal is to predict the long-term task demonstrated in the video (e.g., cooking breakfast, cleaning house, etc). We evaluate a few variants of SlowFast and TimeSformer on this task. ``Single Clip Coverage'' denotes the number of seconds spanned by a single clip. ``\# Test Clips'' is the average number of clips needed to cover the entire video during inference. All models in this comparison are pretrained on Kinetics-400.\vspace{-0.3cm}}
\label{how2_results_table}
\end{table}

We present our results in Table~\ref{how2_results_table}. As our baselines, we use four variants of SlowFast R101, all operating on video clips sampled at a frame rate of $1/32$ but having varying number of frames: $8, 32, 64$ and $96$. We use the same four configurations for TimeSformer, starting from a ViT pretrained on ImageNet-21K. All models in this comparison are pretrained on Kinetics-400 before finetuning on HowTo100M.

During inference, for each method, we sample as many non-overlapping temporal clips as needed to cover the full temporal extent of a video, e.g., if a single clip spans $8.5$ seconds, we would sample $48$ test clips to cover a video of $410$ seconds. Video-level classification is done by averaging the clip predictions.



From the results in Table~\ref{how2_results_table} we first note that, for the same single clip coverage, TimeSformer outperforms the corresponding SlowFast by a large margin of $8-11\%$. We also observe that longer-range TimeSformers do better, i.e., our longest-range variant achieves the best video-level classification accuracy. These results suggest that our model is highly suitable for tasks that require long-term video modeling. 

We also experimented with finetuning TimeSformer directly from a ViT pretrained on ImageNet-1K and ImageNet-21K (skipping the Kinetics-400 training). We report that when pretrained only on ImageNet-1K, our model achieves top-1 accuracies of $52.8, 58.4, 59.2, 59.4$ for $8, 32, 64, 96$ frame inputs, respectively. When considering ImagNet-21K pretraining, TimeSformer produces  top-1 accuracies of $56.0, 59.2, 60.2, 62.1$ for $8, 32, 64, 96$ frame inputs, respectively. These results demonstrate that our model can effectively exploit long-range temporal dependencies regardless of the pretraining dataset that we use.

\begin{figure}
\begin{center}
   \includegraphics[width=1\linewidth]{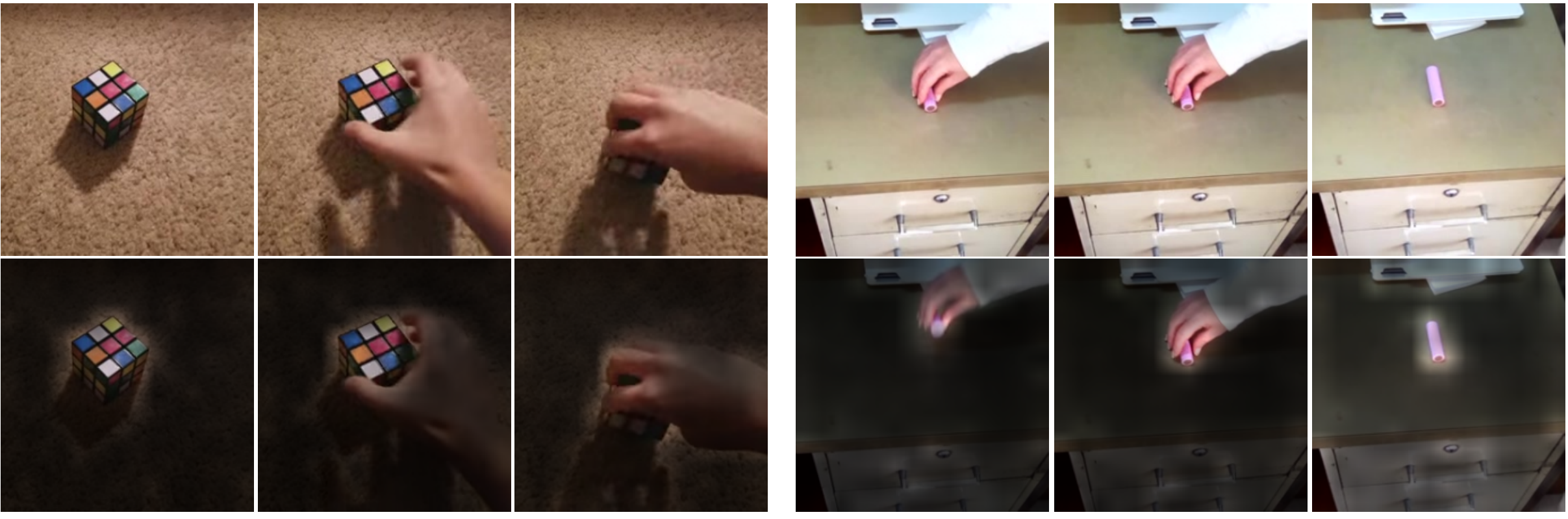}
\end{center}
\vspace{-0.4cm}
        \caption{Visualization of space-time attention from the output token to the input space on Something-Something-V2. Our model learns to focus on the relevant parts in the video in order to perform spatiotemporal reasoning.\vspace{-0.2cm}} 
\label{attention_ex_fig}
\end{figure}

\begin{figure}
\begin{center}
   \includegraphics[width=0.92\linewidth]{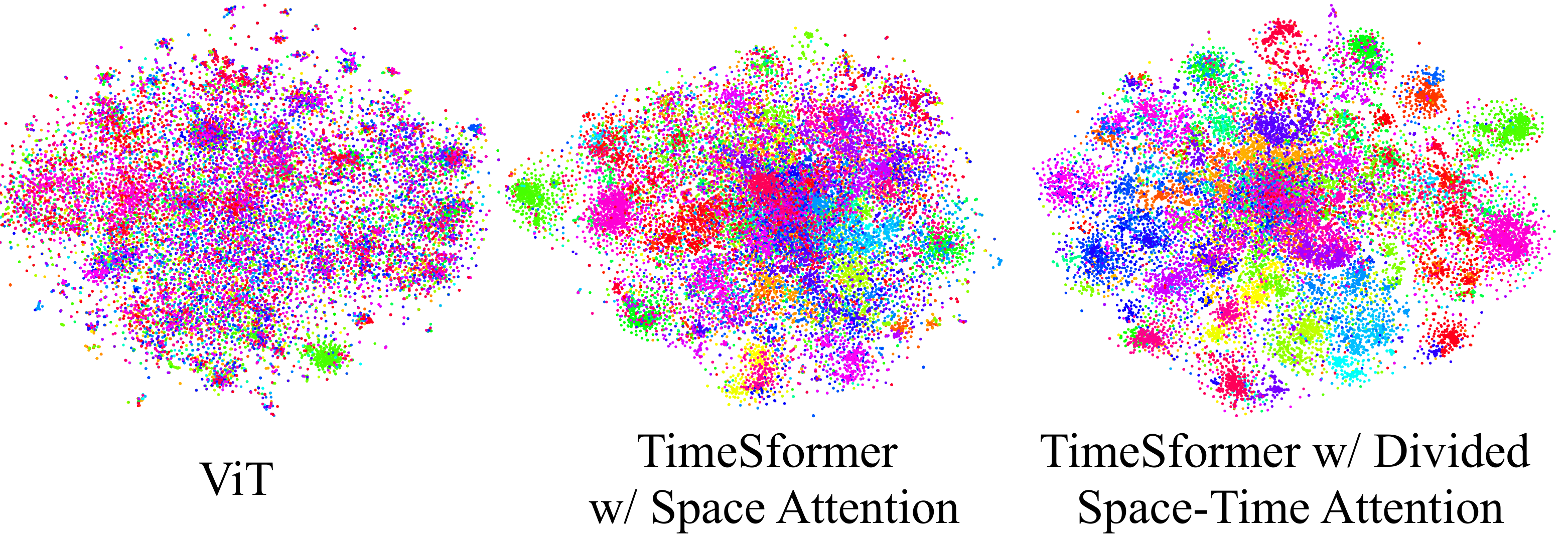}
\end{center}
\vspace{-0.4cm}
        \caption{Feature visualization with t-SNE~\cite{vanDerMaaten2008} on Something-Something-V2. Each video is visualized as a point. Videos belonging to the same action category have the same color. The TimeSformer with divided space-time attention learns semantically more separable features than the TimeSformer with space-only attention or ViT~\cite{Dosovitskiy:ICLR2021}. \vspace{-0.3cm}}
\label{tsne_fig}
\end{figure}

\subsection{Additional Ablations}

\textbf{Smaller \& Larger Transformers.} In addition to the ``Base'' ViT model~\cite{Dosovitskiy:ICLR2021}, we also experimented with the ``Large'' ViT. We report that this yielded results $1\%$ worse on both Kinetics-400, and Something-Something-V2. Given that our ``Base'' model already has $121M$ parameters, we suspect that the current datasets are not big enough to justify a further increase in model capacity. We also tried  the ``Small'' ViT variant, which produced accuracies about $5\%$ worse than our default ``Base'' ViT model.

\textbf{Larger Patch Size.} We also experimented with a different patch size, i.e., $P=32$. We report that this variant of our model produced results about $3\%$ worse  than our default variant using $P=16$.
We conjecture that the performance decrease with $P=32$ is due to the reduced spatial granularity.
 We did not train any models with $P$ values lower than 16 as those models have a much higher computational cost.

\textbf{The Order of Space and Time Self-Attention.} Our proposed ``Divided Space-Time Attention'' scheme applies temporal attention and spatial attention one after the other. Here, we investigate whether reversing the order of time-space attention (i.e., applying spatial attention first, then temporal) has an impact on our results. We report that applying spatial attention first, followed by temporal attention leads to a $0.5\%$ drop in accuracy on both Kinetics-400, and Something-Something-V2. We also tried a parallel space-time self-attention. We report that it produces $0.4\%$ lower accuracy compared to our adopted ``Divided Space-Time Attention'' scheme.

\subsection{Qualitative Results}

\textbf{Visualizing Learned Space-Time Attention.} In Figure~\ref{attention_ex_fig}, we present space-time attention visualizations obtained by applying TimeSformer on Something-Something-V2 videos. To visualize the learned attention, we use the Attention Rollout scheme presented in~\cite{abnar2020quantifying}.  Our results suggest that TimeSformer learns to attend to the relevant regions in the video in order to perform complex spatiotemporal reasoning. For example, we can observe that the model focuses on the configuration of the hand when visible and the object-only when not visible. 

\textbf{Visualizing Learned Feature Embeddings.} In Figure~\ref{tsne_fig}, we also visualize the features learned by TimeSformer on Something-Something-V2. The visualization is done using t-SNE~\cite{vanDerMaaten2008} where each point represents a single video, and different colors depict different action categories. Based on this illustration, we observe that TimeSformer with divided space-time attention learns semantically more separable features than the TimeSformer with space-only attention or ViT~\cite{Dosovitskiy:ICLR2021}.

\section{Conclusion}

In this work, we introduced TimeSformer, a fundamentally different approach to video modeling compared to the established paradigm of convolution-based video networks. We showed that it is possible to design an effective, and scalable video architecture built exclusively on space-time self-attention.  Our method (1) is conceptually simple, (2) achieves state-of-the-art results on major action recognition benchmarks, (3) has low training and inference cost, and (4) can be applied to clips of over one minute, thus enabling long-term video modeling. In the future, we plan to extend our method to other video analysis tasks such as action localization, video captioning and question-answering.



\appendix
\section*{Appendix}

\setcounter{figure}{0}
\setcounter{table}{0}
\renewcommand{\thetable}{A.\arabic{table}}
\renewcommand{\thefigure}{A.\arabic{figure}}

\section{Implementation Details}

Our TimeSformer implementation is built using PySlowFast~\cite{fan2020pyslowfast} and pytorch-image-models~\cite{rw2019timm} packages. Below, we describe specific implementation details regarding the training and inference procedures of our model.

\textbf{Training.} We train our model for $15$ epochs with an initial learning rate of $0.005$, which is divided by $10$ at epochs $11,$ and $14$. During training, we first resize the shorter side of the video to a random value in $[256, 320]$. We then randomly sample a $224 \times 224$ crop from the resized video. For our high-resolution model, TimeSformer-HR, we resize the shorter side of the video to a random value in $[448, 512]$, and then randomly sample a $448 \times 448$ crop. We randomly sample clips from the full-length videos with a frame rate of $1/32$. The batch size is set to $16$. We train all our models using synchronized SGD across $32$ GPUs. The momentum is set to $0.9$, while the weight decay is set to $0.0001$. 

Unless otherwise noted, in our experiments we use the ``Base'' ViT model~\cite{Dosovitskiy:ICLR2021}.  Temporal and spatial attention layers in each block are initialized with the same weights, which are obtained from the corresponding attention layer in ViT. 

\textbf{Inference.} As discussed in the main draft, during inference we sample a single temporal clip in the middle of the video. We scale the shorter spatial side of a video to $224$ pixels (or $448$ for TimeSformer-HR) and take $3$ crops of size $224\times224$ ($448\times448$ for TimeSformer-HR) to cover a larger spatial extent within the clip. The final prediction is obtained by averaging the softmax scores of these $3$ predictions.

\textbf{Other models in our comparison.} To train I3D~\cite{DBLP:conf/cvpr/CarreiraZ17}, and SlowFast~\cite{slowfast}, we use the training protocols that were used in the original papers. For I3D, we initialize it with a 2D ImageNet CNN, and then train it for $118$ epochs with a base learning rate of $0.01$, which is divided by $10$ at epochs $44$ and $88$. We use synchronized SGD across $32$ GPUs following the linear scaling recipe of~\citet{goyal2017imagenet1hr}. We set the momentum to $0.9$, and weight decay to $0.0001$. The batch size is set to $64$. For the SlowFast model, when initialized from ImageNet weights, we use this same exact training protocol. When training SlowFast from scratch, we use the training protocol described by the authors~\cite{slowfast}. More specifically, in that case, the training is done for $196$ epochs with a cosine learning rate schedule, and the initial learning rate is set to $0.1$. We use a linear warm-up for the first $34$ epochs starting with a learning rate of $0.01$. A dropout of $0.5$ is used before the final classification layer. The momentum is set to $0.9$, the weight decay is $0.0001$, and the batch size is set to $64$. Just as before, we adopt the linear scaling recipe~\cite{goyal2017imagenet1hr}.

\textbf{Datasets.} Kinetics-400~\cite{DBLP:conf/cvpr/CarreiraZ17} consists of $240K$ training
videos and $20K$ validation videos that span $400$ human action categories. Kinetics-600~\cite{DBLP:journals/corr/abs-1808-01340} has  $392K$ training videos and $30K$ validation videos spanning $600$ action categories. Something-Something-V2~\cite{DBLP:journals/corr/GoyalKMMWKHFYMH17} contains $170K$ training videos and $25K$ validation videos that span $174$ action categories. Lastly, Diving-48~\cite{Li_2018_ECCV} has $16K$ training videos and $3K$ testing videos spanning $48$ fine-grained diving categories. For all of these datasets, we use standard classification accuracy as our main performance metric.

\bibliography{gb_bibliography}
\bibliographystyle{icml2021}

\end{document}